\documentclass{article}

\usepackage{arxiv}

\usepackage[utf8]{inputenc} 
\usepackage[T1]{fontenc}    
\usepackage{hyperref}       
\usepackage{url}            
\usepackage{booktabs}       
\usepackage{amsfonts}       
\usepackage{nicefrac}       
\usepackage{microtype}      
\usepackage{lipsum}
\usepackage{parskip}
\usepackage{graphicx}
\usepackage{float}
\graphicspath{{plots/}}
\usepackage{caption}
\usepackage{bm}
\usepackage[numbers, super]{natbib}
\usepackage[switch]{lineno}

\title{Bayesian deep learning for mapping via auxiliary information: a new era for geostatistics?}

\author{
  Charlie Kirkwood\\
  Department of Mathematics\\
  University of Exeter, UK\\
  \texttt{c.kirkwood@exeter.ac.uk} \\
  \And
  Theo Economou\\
  Department of Mathematics\\
  University of Exeter, UK\\
    \And
  Nicolas Pugeault\\
  School of Computing Science\\
  University of Glasgow, UK\\
}

\begin{document}
\maketitle

\begin{abstract}

Earth scientists increasingly deal with `big data'. For applications involving spatial modelling and mapping, variants of kriging --- the spatial interpolation technique developed by South African mining engineer Danie Krige --- have long been regarded as the established geostatistical methods. However, kriging and its variants (such as regression kriging, in which auxiliary variables or derivatives of these are included as covariates) are relatively restrictive models and lack capabilities that have been afforded to us in the last decade or so by deep neural networks. Principal among these is feature learning: the ability to learn filters to recognise task-specific patterns in gridded data such as images. Here we demonstrate the power of feature learning in a geostatistical context, by showing how deep neural networks can automatically learn the complex relationships between point-sampled target variables and gridded auxiliary variables (such as those provided by remote sensing), and in doing so produce detailed maps of chosen target variables. At the same time, in order to cater for the needs of decision makers who require well-calibrated probabilities, we demonstrate how both aleatoric and epistemic uncertainty estimates can be obtained from deep neural networks via a Bayesian approximation known as Monte Carlo dropout. In our example, we produce a national-scale probabilistic geochemical map from point-sampled observations, with auxiliary data provided by a terrain elevation grid. Unlike traditional geostatistical approaches, auxiliary variable grids are fed into our deep neural network raw. There is no need to provide derivatives (e.g. slope angles, roughness - in the case of terrain) because the deep neural network is capable of learning these and arbitrarily more complex derivatives as necessary to optimise prediction. We hope our positive results will raise awareness of the suitability of Bayesian deep learning --- and its feature learning capabilities --- for large-scale geostatistical applications where uncertainty matters.
\end{abstract}


\twocolumn

\begin{figure*}[!htb]
    \centering
    \includegraphics[width=\textwidth]{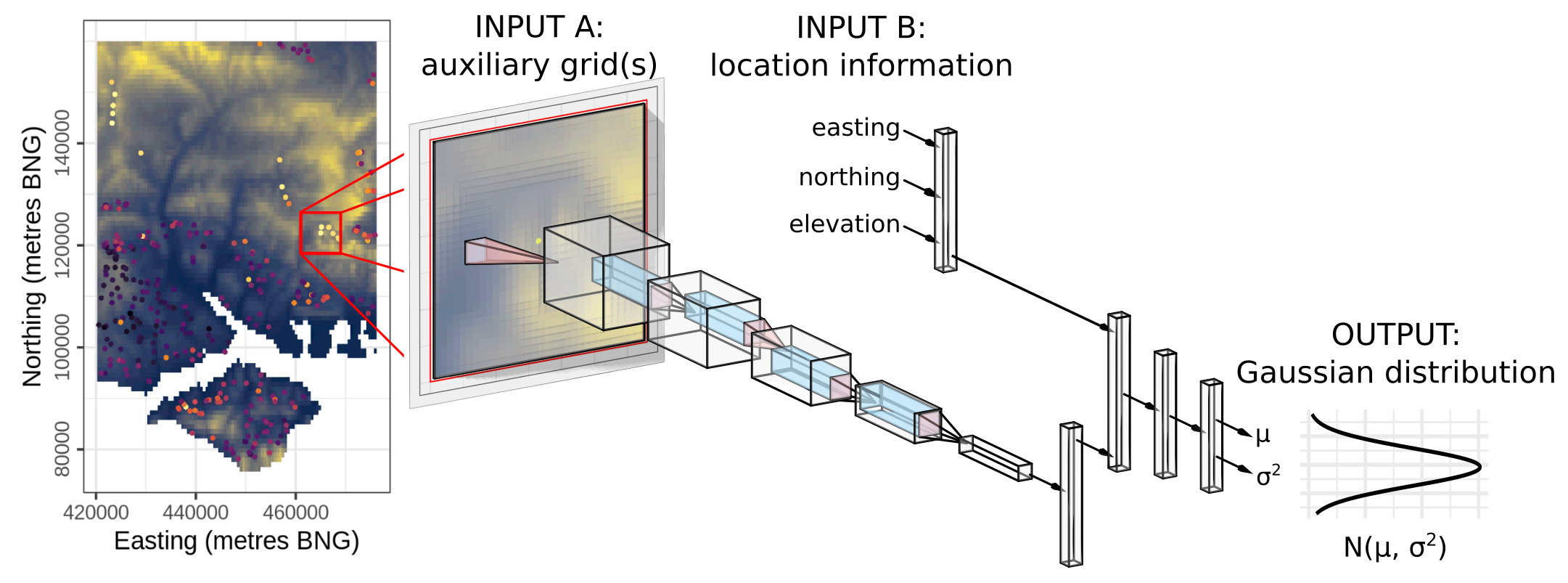}
    \caption{Overview of our geostatistical deep neural network. For each point-sampled target variable observation we provide the neural network with an image of surrounding terrain (our auxiliary grid in this case) centred on the observation site. We also provide location information as a secondary input. The neural network optimises information extraction from these inputs in order to make probabilistic predictions about the value of the target variable. Once trained, the neural network can make predictions for any point in space, by providing the corresponding terrain image and location information.}
    \label{fig:overview}
\end{figure*}

Maps are important for our understanding of Earth and its processes, but it is generally the case that we are unable to directly observe the variables we are interested in at every point in space. For this reasons we must use models to fill in the gaps. In order to support decision making under uncertainty, statistical models are desirable\cite{berger1985statistical}. Kriging --- the original geostatistical model\cite{cressie1990origins} --- provides smooth interpolations between point observations based on the spatial autocorrelation of a target variable\cite{stein1999interpolation}. However, additional sources of information are often available. The rise of remote sensing\cite{mulder2011use, colomina2014unmanned} has provided readily accessible grids of what we consider here to be auxiliary variables (e.g. terrain elevation, spectral imagery, subsurface geophysics). These are complete maps of variables that we are not directly interested in but which are likely to contain information relating to our variables of interest. How best to extract such information from auxiliary variable grids for geostatistical modelling tasks has remained an open question, but often involves trial-and-error experimentation using manually designed filters to extract features with as much explanatory power as possible\cite{ruiz2011topographic, poggio2013regional, parmentier2014assessment, kirkwood2016machine, kirkwood2016dropout, young2018spatial, lamichhane2019digital} (e.g. deriving slope angle as a feature to explain landslide susceptibility\cite{youssef2016landslide}). Here we present an end-to-end geostatistical modelling framework using Bayesian deep learning, which frames the information extraction problem as an optimisation problem\cite{shwartz2017opening}, and in doing so eliminates the need for manual feature engineering and feature selection steps.

Our two-branch deep neural network architecture --- convolutional layers for feature learning combined with dense layers for smooth interpolation --- brings the benefits of deep learning to geostatistical applications, and we do so without sacrificing uncertainty estimation: Our approach estimates both aleatoric and epistemic uncertainties (via Monta Carlo dropout\cite{gal2016dropout}) in order to provide a theoretically grounded predictive distribution as output. Our work brings together ideas from the fields of machine learning\cite{krizhevsky2012imagenet, srivastava2014dropout}, remote sensing\cite{zhang2016deep, zhu2017deep} and Bayesian geostatistics\cite{handcock1993bayesian, pilz2008we}, and unites them in a general framework for solving `big data' geostatistical modelling tasks in which gridded auxiliary variables are available to support the interpolation of point-sampled target variables. As far as we are aware, our framework is the first to provide both well-calibrated probabilistic output and automated feature learning in the context of geospatial regression tasks. By using neural networks, we also ensure that our framework is scalable to the largest of problems.

While the framework we present here is new, it can also be seen as a unification and generalisation of a range of prior works. Deep learning\cite{lecun2015deep} --- machine learning using deep neural networks --- has seen increasing uptake within scientific communities in the last decade, after breakthrough work by \citet{krizhevsky2012imagenet} who achieved a new state of the art in image classification by using deep neural networks to automatically learn informative features (rather than manually derive them). Deep learning has been widely adopted within the remote sensing community\cite{zhang2016deep, zhu2017deep, li2017estimating, zuo2019deep}. However, difficulty in obtaining reliable uncertainty estimates from deep neural networks\cite{kendall2017uncertainties} has meant that deep learning has not been widely adopted for applications where uncertainty matters. A few authors have applied feature learning in a geostatistical context\cite{padarian2019using, wadoux2019multi, wadoux2019using, kirkwood2020deep} (mostly for digital soil mapping), but only one --- \citet{wadoux2019using} --- has been able to provide uncertainty estimates (though these were achieved via a bootstrapping approach and found to be underdispersive). Here we make use of a theoretically grounded and practically effective approach to uncertainty estimation: Monte Carlo dropout as a Bayesian approximation, as conceived by \citet{gal2016dropout}. The authors are aware of one prior instance of its use within remote sensing: for a semantic segmentation task by \citet{kampffmeyer2016semantic}. Therefore, while separate concepts behind our work may be familiar to some readers already, we believe the time is right to bring them together and present Bayesian deep learning as a general solution for big data geostatistics, with our demonstrated methodology that we hope will become widely used in future.

\section*{Feature learning, for geostatistics}

The core domain of geostatistics has been in the spatial interpolation of point observations in order to produce continuous maps in two or three dimensions. Kriging, the now ubiquitous geostatistical technique conceived by South African mining engineer Danie Krige\cite{krige1951statistical}, originally accounted for only the location and spatial autocorrelation of observations in order to produce smooth interpolations that can be considered optimal if no other information is available\cite{matheron1962traite}. When other information is available, as is commonly the case today thanks to remote sensing, the pursuit of optimal spatial interpolation becomes more complex. An extension of ordinary kriging, regression kriging, allows covariates to be included in the model: the mean of the interpolated output is able to vary as a linear function of the value of covariates at the corresponding location\cite{gotway1996geostatistical, hengl2007regression}. For an illustrative example, the inclusion of elevation as a covariate in an interpolation of surface air temperature data could be expected to result in a map that reflects the underlying elevation map, i.e. whose mean function is a linear function of elevation. However, this quickly brings us to the limits of regression kriging: what if a linear function of elevation does not provide as much explanatory power to surface air temperature as some non-linear function of elevation? What non-linear function of elevation would be the optimal one? At the same time, what if we also have wind direction available to use as a covariate. Wouldn't the best predictor of surface air temperature account for not just elevation, or wind direction, but how they interact, with air flowing down from mountains expected to be cooler? We quickly find ourselves in the realms of feature engineering and feature selection, a world of hypothesising and trial-and-error experimentation which has become a necessary but impractical step in the traditional geostatistical modelling process.

\begin{figure}[!htb]
    \centering
    \includegraphics[width=\columnwidth]{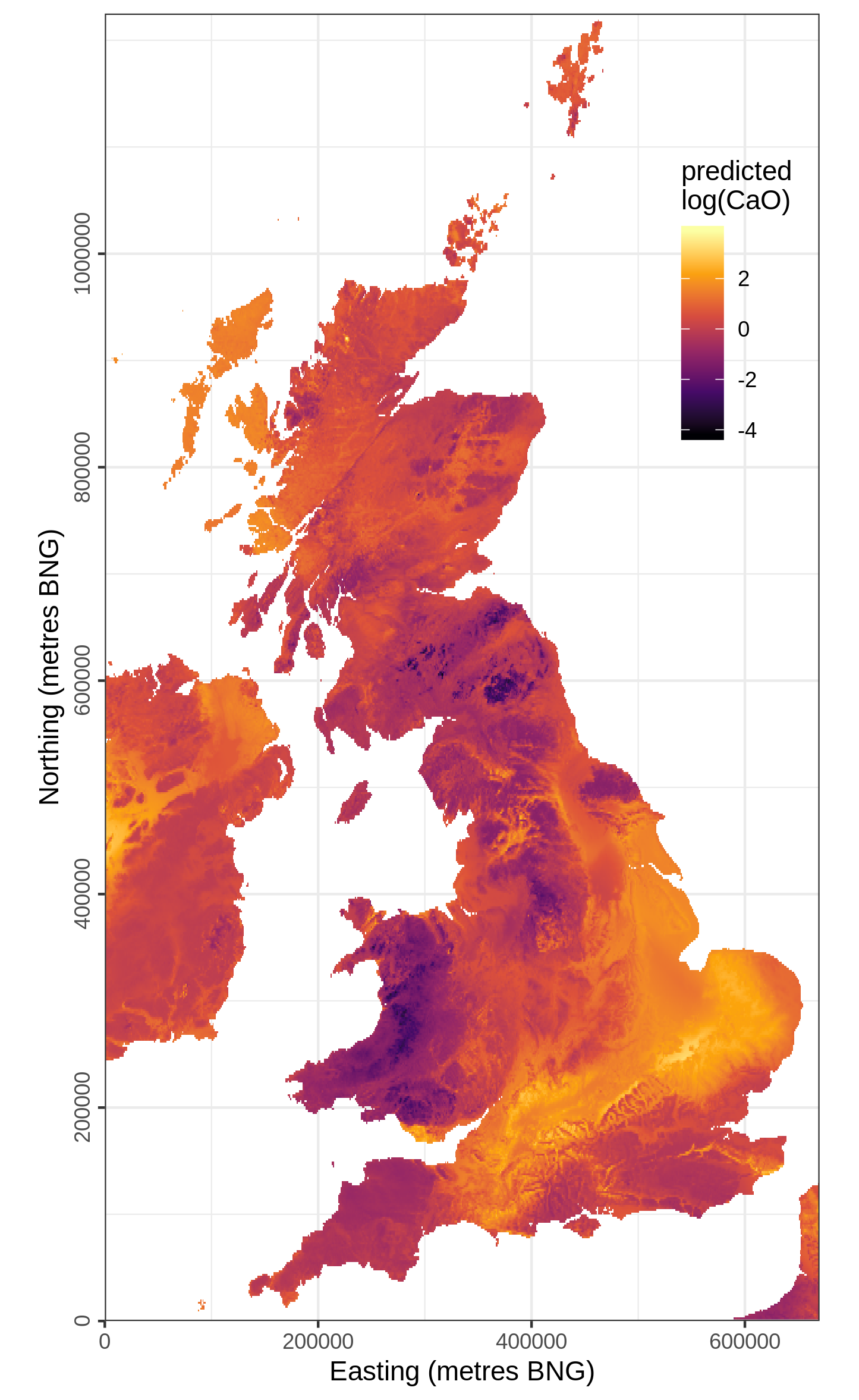}
    \caption{Deep learned map of log(CaO) in stream sediments across the UK. This deterministic map uses the mean of the predictive distribution.}
    \label{fig:national}
\end{figure}

The defining strength of deep neural networks is their ability to learn features for themselves owing to their hierarchical structure in which the output of each layer (multiplied by a non-linear activation function) provides the input to the next. Through back-propagation of error gradients, neural networks can automatically learn non-linear transformations of input variables and their interactions as necessary in order to minimise loss. It has also been shown that in the limit of infinite width (infinite number of nodes) a neural network layer becomes mathematically equivalent to a Gaussian Process\cite{neal1996priors}, which is itself the same smooth interpolator conceived by Danie Krige under a different name. It is not surprising then that neural networks are very capable spatial interpolators. However, our deep neural network combines these spatial abilities with a unique ability to learn its own features from auxiliary variable grids. We achieve this efficiently through the use of convolutional layers --- trainable filters which pass over gridded data to derive new features, in a similar manner to how edge detection filters derive edges from photographs \cite{chen2017hyperspectral}. By stacking convolutional layers, the complexity and scale of features that can be derived increases, along with the size of the receptive field of the neural network\cite{luo2016understanding}, which allows longer-range dependence structures to be learned. In our example, we use elevation data from NASA's Shuttle Radar Topography Mission\cite{van2001shuttle} as our gridded auxiliary variable. Our target variable is the (log transformed) concentration of calcium in stream sediments, collected by the British Geological Survey\cite{johnson2005g}. We chose calcium for its relatively high mobility, and therefore complex dependence on terrain topography, the learning of which provides a suitable challenge for our deep neural network's feature-learning capabilities.

\section*{Estimating uncertainties via MC dropout}

\begin{figure*}
    \centering
    \includegraphics[width=0.33\textwidth]{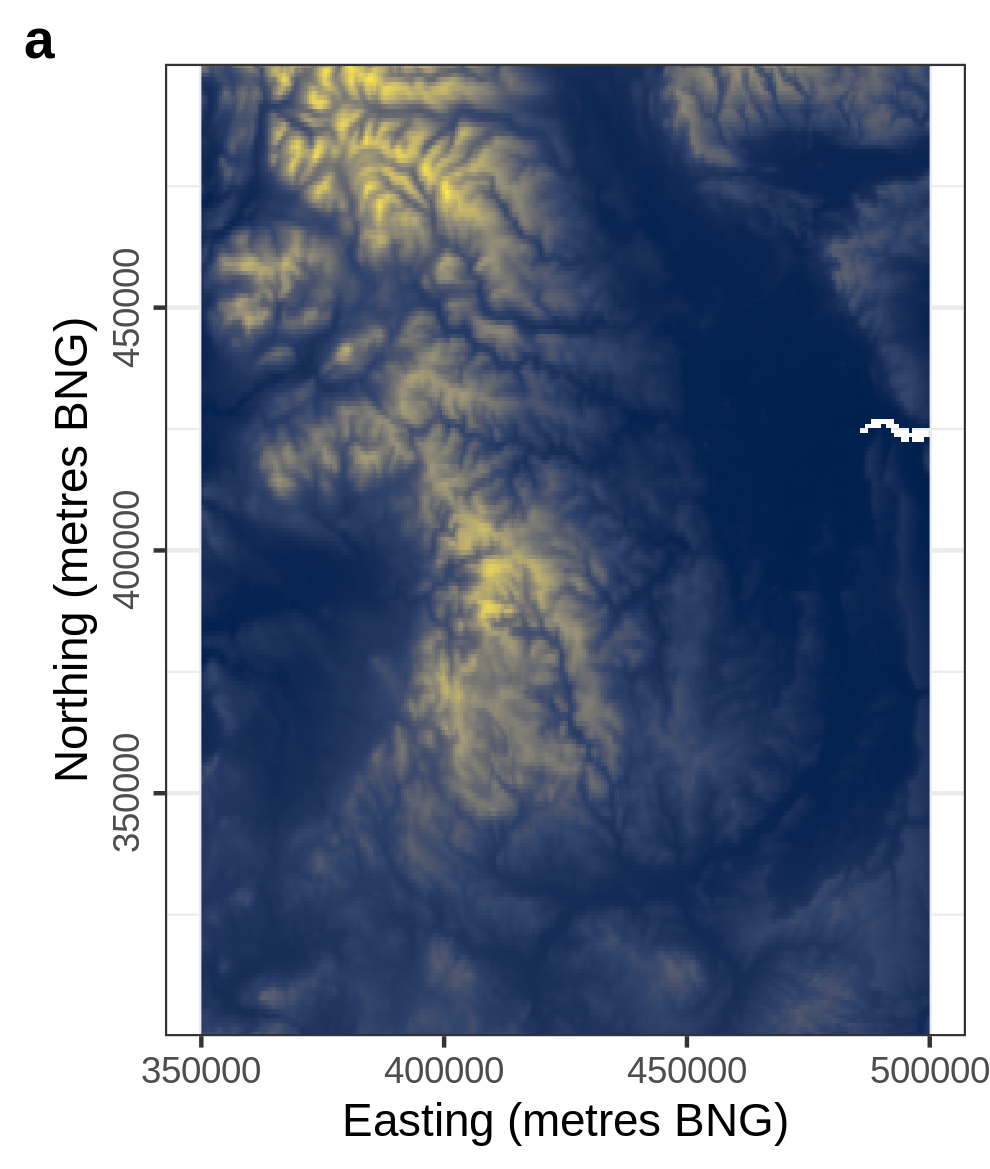}
    \includegraphics[width=0.33\textwidth]{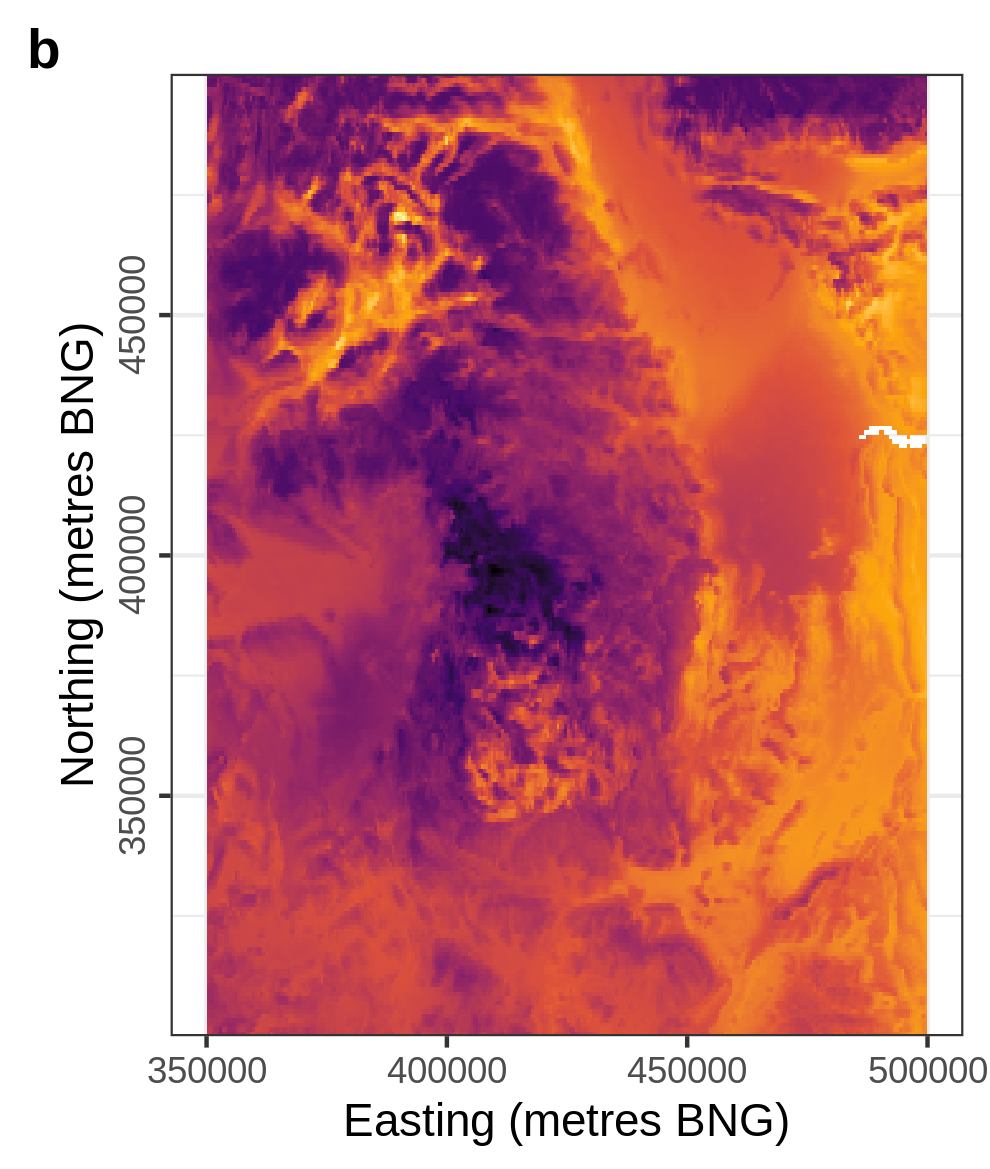}
    \includegraphics[width=0.33\textwidth]{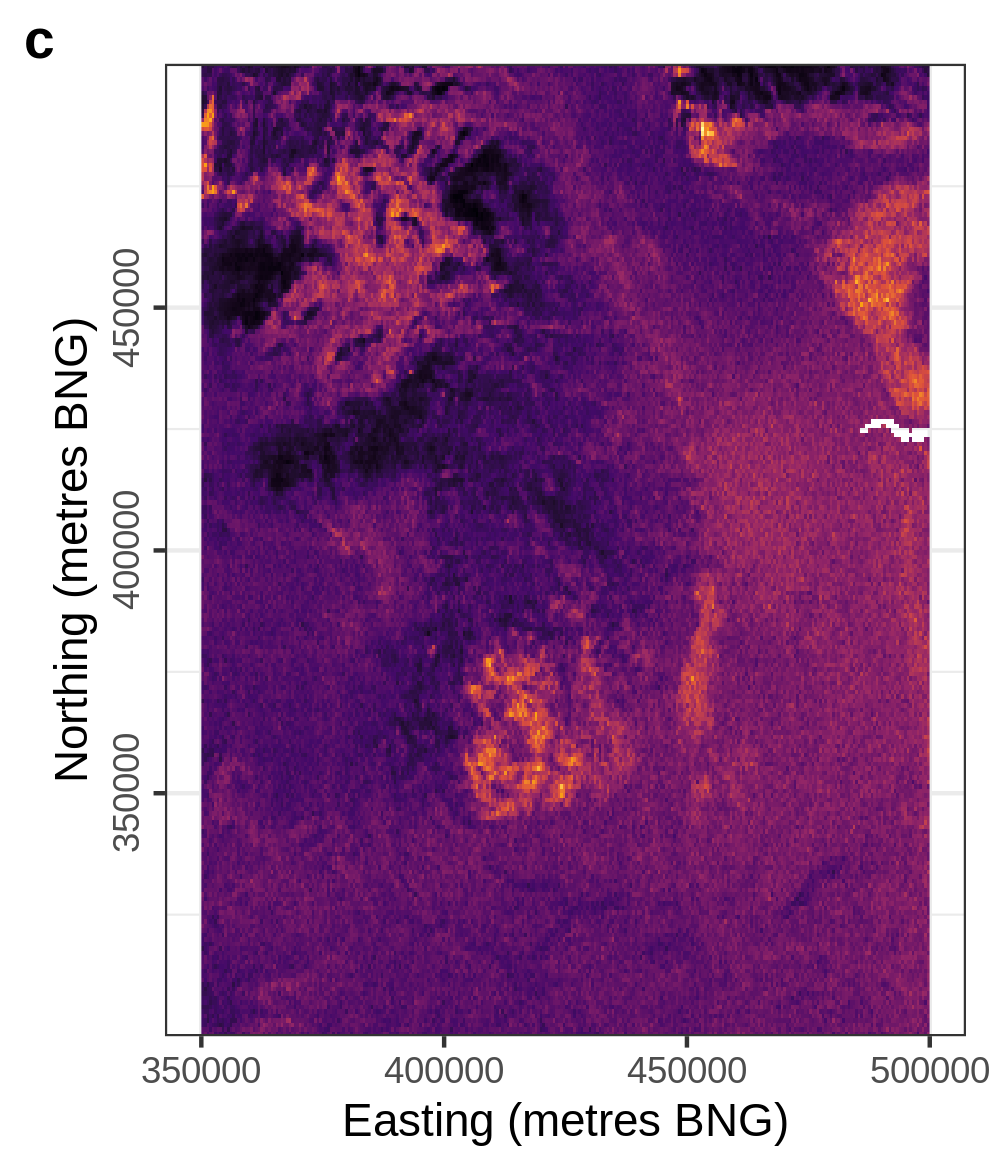}
    \caption{Zooming in on a local area: \textbf{a}, SRTM elevation data: our gridded auxiliary variable. \textbf{b}, Predicted log(CaO): deterministic mean. \textbf{c}, Uncertainty of predicted log(CaO): standard deviation of posterior predictive distribution. All maps use linear colour scales where brighter = greater. The white inlet is the tip of the Humber estuary.}
    \label{fig:locals}
\end{figure*}

\begin{figure*}[!htb]
    \centering
    \includegraphics[width=\textwidth]{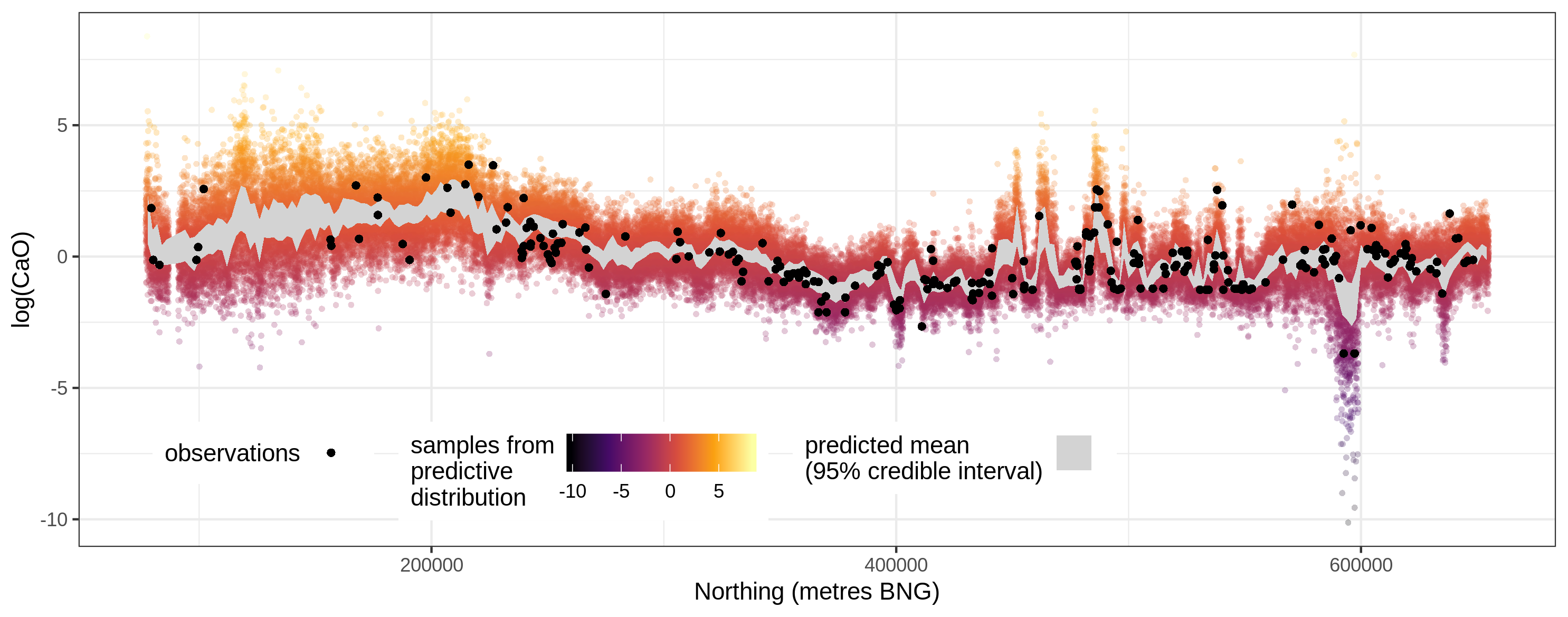}
    \caption{South-north cross section of our Bayesian deep neural network output, running along a line at 400000 metres easting BNG. Also shown are all the observations within 500m either side of this line.}
    \label{fig:xsection}
\end{figure*}

Neural networks tend to be deterministic: trained through back-propagation to converge on a set of weights that minimise a loss function. In most neural networks these final weights are fixed, having no distribution, which means that there is no way to estimate the uncertainty in these weights or in the overall function (or learned model) that the neural network represents. Natural processes inevitably involve uncertainties, and it is right that we should want to quantify these in order to provide well-calibrated probabilistic predictions suitable for use in decision support\cite{yoe2011principles, fox2011distinguishing}. Aleatoric uncertainty, the uncertainty arising from natural randomness in a process, can be accounted for by using a parametric distribution as the output of the neural network. In our case, our deep neural network outputs two parameters - the mean ($\mu$) and variance ($\sigma^2$) of a Gaussian distribution. However, this output distribution alone does not help us to estimate the uncertainty in the form of the model itself --- epistemic uncertainty. To do so requires a distribution over the neural network weights, and for this we use Monte Carlo dropout as proposed by Yarin Gal\cite{gal2016dropout}. This approach places a Bernoulli prior over the weights, which means that for every iteration of training and prediction, the nodes of the neural network each have a probability of being switched off, or dropped out. The probability or rate at which nodes will drop out is a tuneable hyper-parameter. While a Bernoulli prior may seem `unrealistic' for individual parameters --- why should a parameter only exist with a fixed probability? --- the overall effect of Monte Carlo dropout on the network as a whole is to turn our single neural network into an almost infinite self-contained ensemble. Each different configuration of dropped nodes realises a different function (or model) from the ensemble. The dropout rate relates to the variance we expect to see between different functions drawn from the ensemble - it acts as a prior over functions. Even with the constraint of a fixed dropout rate, the neural network is able to locally adjust its epistemic uncertainty during the training process. This, in combination with manual tuning of the dropout rate (to minimise loss on an evaluation set), allows our neural network to produce a well-calibrated posterior predictive distribution suitable for decision support.

\section*{Application to geochemical mapping}

We applied our Bayesian deep neural network to the task of mapping stream sediment calcium concentrations, as log(calcium oxide), across the UK. The dataset contains 109201 point-sampled calcium observations\cite{johnson2005g}, which we split at random into 10 folds, of which one was set aside as a final test dataset (from which we report our prediction accuracy and calibration results), one was used as an evaluation set during the neural network training (to monitor loss on out-of-sample data, to guide hyper-parameter tuning), and the remaining eight folds were used as the training set. We use NASA's Shuttle Radar Topography Mission elevation data\cite{van2001shuttle} as our gridded auxiliary variable. We appended each calcium observation with a 32x32 cell image of the surrounding terrain centred on the observation. We use a grid cell size of 250m, which means that the neural network has an 8x8km square window centred on each observation from which to learn its terrain features. In order to facilitate the learning of terrain features, we normalise each 32x32 cell image so that the centre point is at zero elevation. Features are then learned purely in relation to the sample site, rather than to absolute elevation. However absolute elevation, along with easting and northing, are provided explicitly as the second input to the neural network (after the convolutional layers - see \autoref{fig:overview}) in order to provide the network with awareness of overall location in the geographic space as well as awareness of local topography. We used early stopping to train our neural network a suitable number of epochs by stopping when predictive performance on the evaluation set was no longer improving (indicating that our posterior predictive distribution had converged on a good approximation of the true data distribution). This tended to occur at around 250 epochs. Note that all results are reported on the third dataset --- the test dataset --- which was not used during training at all.

\section*{Results and discussion}

\begin{figure}[!htb]
    \centering
    \includegraphics[width=\columnwidth]{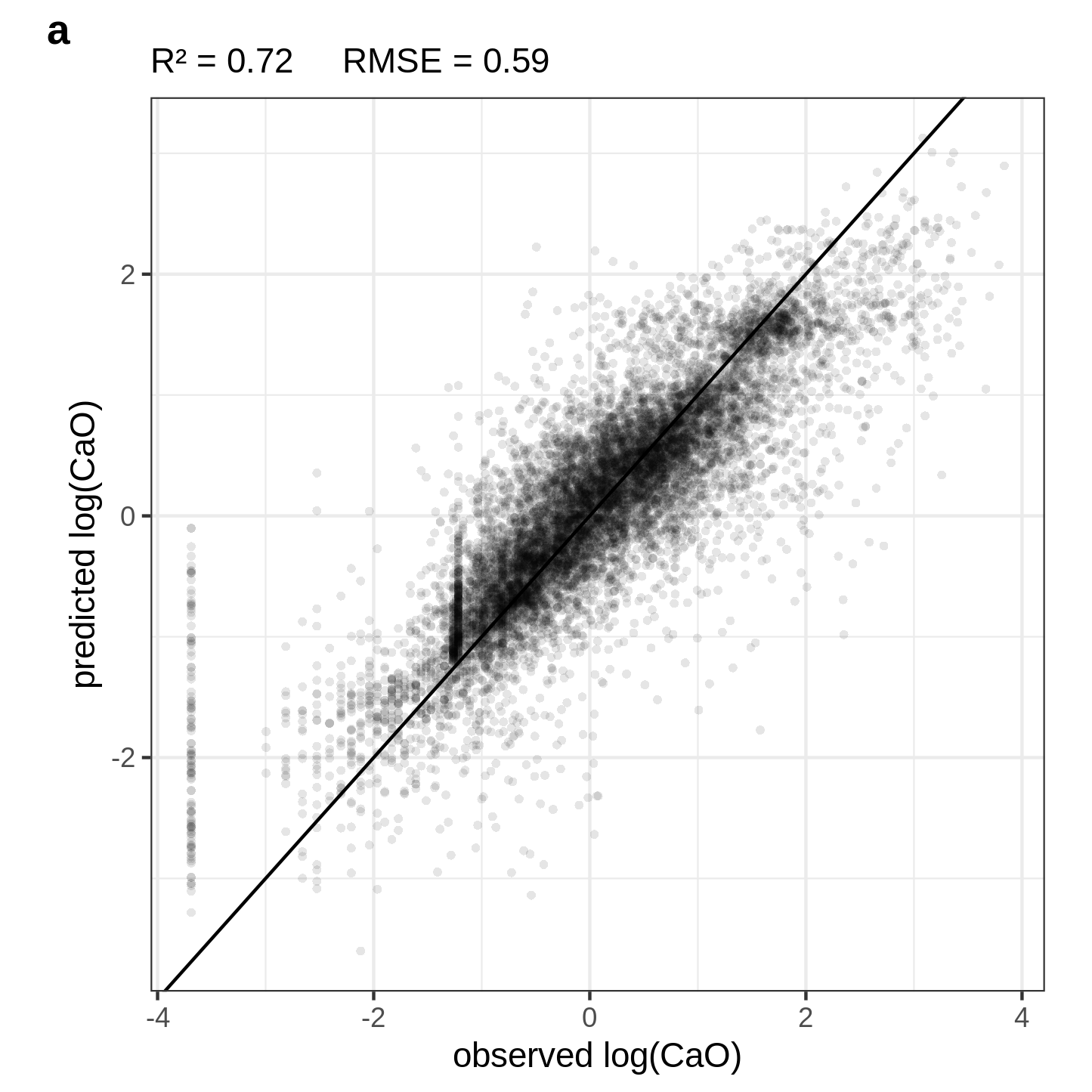}
    \includegraphics[width=\columnwidth]{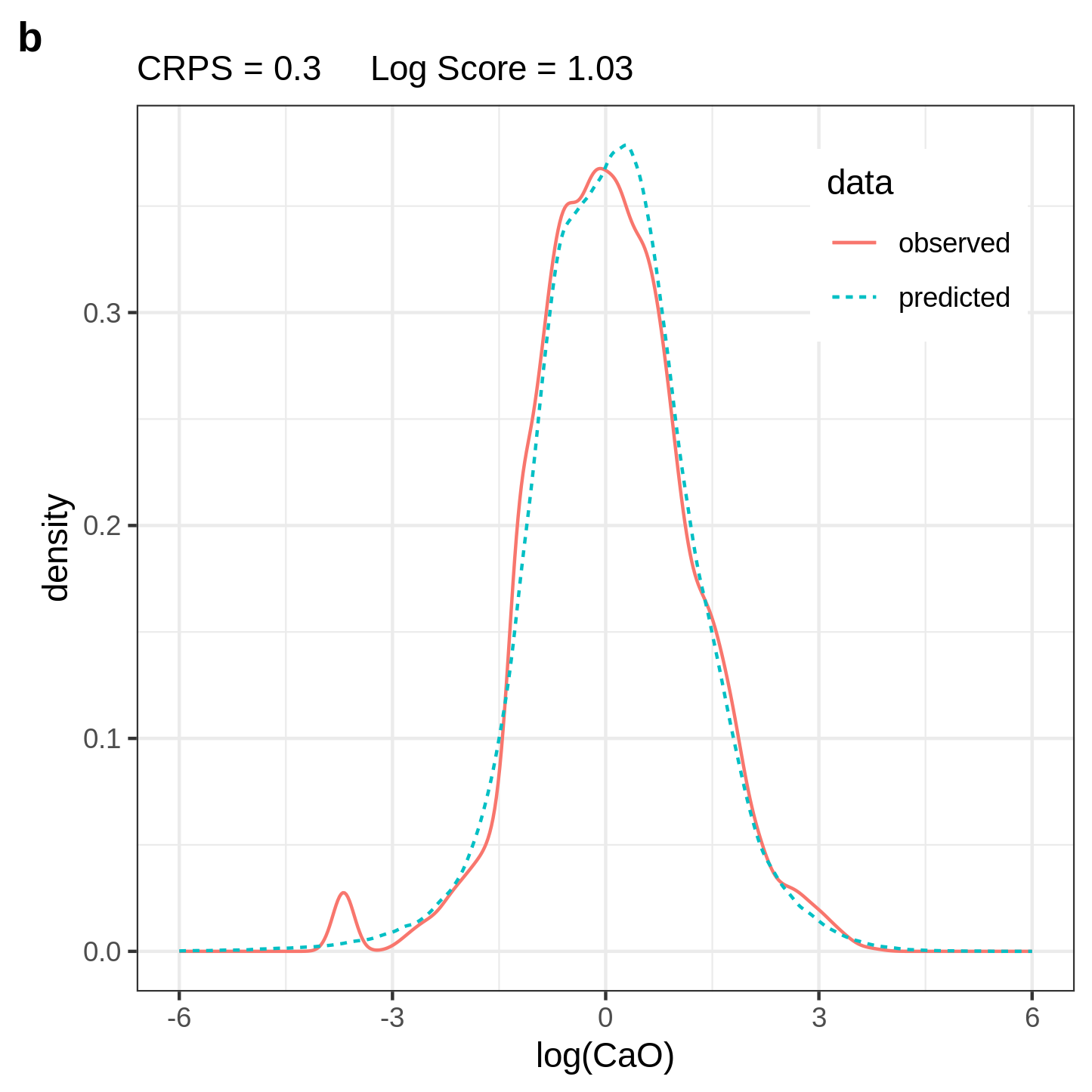}
    \caption{\textbf{a}, Comparison of observed and predicted values of log(CaO) in a deterministic context, taking the mean of the predictive distribution.\textbf{b}, Comparison of observed and predicted distributions of log(CaO) in a probabilistic context, taking 50 samples from the predictive distribution for each observation. Both use data from the held out test set (n = 10920), unseen until after model training and hyper-parameter tuning.}
    \label{fig:distributions}
\end{figure}

The national scale geochemical map that our Bayesian deep neural network has produced is extremely detailed and appears to have successfully captured the complex relationships between our target variable: stream sediment calcium concentrations, and our auxiliary variable grid: terrain elevation (\autoref{fig:national}). Detail alone would be worthless if it did not correspond to good predictive performance, but our results on held out test data --- unseen during the model training and hyper-parameter tuning procedure --- are very encouraging: In a deterministic sense, the mean prediction from our Bayesian deep neural network explains 72\% of the variance in our target variable. The performance of the network in a probabilistic sense is less easily summarised by a single number, but a comparison of the predictive distribution with the true distribution on the held out test set (\autoref{fig:distributions}) indicates a well-calibrated fit\cite{gneiting2007probabilistic}. We have also measured performance using two proper scoring rules\cite{gneiting2007strictly}: the continuous rank probability score (CRPS) and logarithmic score, though these will be most useful in future comparisons with other models. We further interrogate the quality of calibration by assessing the coverage of prediction intervals, and find that 94.9\%, 70.2\%, and 50.1\% of observations fall within the 95\%, 70\%, and 50\% prediction intervals respectively. Such precise coverage, on a relatively large held out test set (10920 observations), is strong evidence that our Bayesian deep neural network is outputting trustworthy probabilities\cite{gneiting2014probabilistic}, and therefore is suitable for supporting decision making under uncertainty.

We visualise the probabilistic capabilities of our deep neural network using a south-north section line through the map (\autoref{fig:xsection}). In doing so, we can see that the neural network is able to capture epistemic and aleatoric uncertainty independently as necessary to minimise loss. The credible interval for the mean varies spatially despite the fixed rate of Monte Carlo dropout, showing that the neural network is able to capture spatial variability in epistemic uncertainty. Likewise, the aleatoric uncertainty also varies spatially as necessary, and is greatest just south of 600000 metres northing, despite low epistemic uncertainty in that region. By outputting a full predictive distribution, the Bayesian deep learning approach can provide probabilistic answers to all sorts of questions\cite{cawley2007predictive, kirkwood2020framework}. Probabilities of exceedance at any location, for example, can be calculated simply as the proportion of probability mass in excess of any chosen threshold.

We zoom in on the national scale map, and visualise predictive uncertainty in \autoref{fig:locals}. Viewing the deterministic mean map at this finer scale, and comparing it to the elevation map of the same extent reveals in yet more detail the remarkable ability of our deep convolutional neural network to learn --- from scratch --- the complex ways in which the distribution of our target variable relates to features of terrain. The same level of complexity is reflected in the uncertainty map, and shows that in addition to being extremely well calibrated (\autoref{fig:distributions}), our Bayesian neural network is also extremely specific and precise in how it models uncertainty, which is desirable.

Our deep neural network is able to produce such specific and detailed outputs because it is interpolating not just in geographic space --- as in traditional geostatistical models --- but also in terrain texture space. This has important implications for mapping tasks. In traditional geostatistical models any predictions made outside the geographic extent of observations would be considered to be extrapolations, and are likely to have high error and uncertainty\cite{journel1989we}. In our case, because our neural network is working in a hybrid space, predictions that would be considered out of sample geographically can still be within sample in terms of terrain features. Deep learning based geostatistical approaches like ours therefore have the potential to make sensible predictions beyond the geographic extent of observations, working on the strength of the relationship between the target variable and the auxiliary variable grids, rather than purely spatial relationships alone. This has implications for optimal sample design in the age of `deep geostatistics', a discussion which we save for future work, other than to say that in this new age, sample design should consider both aspects of the hybrid space: the geographic space and the terrain feature space.

The ability to interpolate in the hybrid space brings with it significant implications for applications like mineral exploration, where obtaining sensible predictions for unexplored regions is a key driver of new discoveries\cite{sabins1999remote}. In our example, the geochemical dataset on which we trained our neural network in fact has no observations for the Republic of Ireland, and yet the predictions for this region (the southern two thirds of the island of Ireland in \autoref{fig:national}) subjectively appear just as plausible as for any other. It would be interesting to evaluate these predictions against Irish geochemical data in future in order to further investigate the abilities of Bayesian deep learning for out-of-region geochemical exploration.

The effects of fluvial processes on calcium are perhaps the most noticeable terrain-related effects captured in the map, with downslope `washing out' of calcium clearly visible. This indicates that our neural network has been able to learn complex physical processes by example. The authors are aware of no other methods that could match the capabilities of our Bayesian deep learning approach in this geochemical mapping task. Numerical models may be able to represent physical processes more accurately, but would be nigh on impossible to parameterise here, and would struggle to accurately quantify uncertainties regardless. Conversely, traditional geostatistical modelling approaches like regression kriging may do well at quantifying uncertainties, but have no capabilities in feature learning. An approach known as topographic kriging\cite{laaha2014spatial} has been developed specifically for interpolation on stream networks, but this is unable to generate predictions outside of the manually designated stream network, and so is of limited use for general mapping applications. We therefore postulate that the Bayesian deep learning architecture we present here represents a real step change in capabilities over previous geostatistical approaches, and we encourage its adoption as a new general solution for `big data' geostatistical problems.

\subsection*{Acknowledgements}
We acknowledge funding from the UK's Engineering and Physical Sciences Research Council (EPSRC project ref: 2071900). Our  thanks go to Nvidia and their GPU grant scheme for kindly providing the Titan X Pascal GPU on which our deep neural network can be trained in under 30 minutes. We also thank the British Geological Survey for making the G-BASE geochemical dataset available for this study.

\subsection*{Data availability}

The code to reproduce this study is available at https://github.com/charliekirkwood/deepgeostat and includes functions to download NASA's SRTM elevation data via the raster package in R. We are however unable to provide open access to our stream sediment geochemistry target variable dataset, however for academic research purposes, readers may request access to this dataset from the British Geological Survey at https://www.bgs.ac.uk/enquiries/home.html or by email to enquiries@bgs.ac.uk.

\bibliographystyle{unsrtnat}  
\bibliography{ref}

\section*{Methods}
\subsection*{Data and preparation}

In our example, we make use of two datasets. For our auxiliary variable grid we use 90m gridded elevation data provided by NASA's Shuttle Radar Topography Mission (SRTM)\cite{van2001shuttle}, which we access via the Raster package in R\cite{robert2017raster,rcoreteam2020r}. In total for the UK this provides 611 404 grid cell elevation values. It is worth noting that our framework is also able to use multiple auxiliary variable grids at once. For our target variable observations, we use stream sediment calcium concentrations (as calcium oxide, CaO) provided by the British Geological Survey's Geochemical Baseline Survey of the Environment (G-BASE) project\cite{johnson2005g}. We work with these concentrations log transformed, such that our target variable is log(CaO), which follows an approximately Gaussian distribution. Excluding NA values, this dataset provides 109 201 stream sediment calcium observations.

Constructing our study dataset required data wrangling in order to link two input types (location information, and an observation-centred terrain image) to each observation of our target variable. Location information is straightforward, and simply consists of the easting and northing values recorded for each observation in the G-BASE dataset, along with an elevation value extracted from the SRTM elevation grid at that location. Creation of the observation-centred terrain image is slightly more complex, and requires extracting 1024 elevation values on a regular 32x32 grid centred at the  location of the observation site. Bicubic interpolation of the elevation grid was used in all elevation extractions, in order to avoid aliasing issues in the terrain images.

Once assembled, the inputs were normalised, as is standard procedure when working with neural networks. For each location variable; easting, northing, and elevation, we simply subtract the mean and divide by the standard deviation, such that they become standardised variables with mean 0 and standard deviation 1. Along the same principal, for the terrain images we subtract the elevation at the observation site (the centre of the image) from the image such that the centre of all terrain images is set to zero. We then divide by the standard deviation of the UK elevation grid so that all terrain images are on the same reduced scale. This normalisation procedure ensures that terrain features learned by the neural network represent context relative to the observation site rather than just recognising differences in absolute elevation (which the explicit elevation variable already provides). We would recommend the same principal for all gridded covariates.

We chose the SRTM and G-BASE datasets for their ease of access and use as well as for the complexity of the spatial relationships they contain, which we believe provide a good demonstration of the capability of our Bayesian deep learning approach for geostatistical modelling tasks. The methodology we present in this paper is intended as a general framework for data-rich geostatistical applications where gridded auxiliary variables are available in addition to point-sampled observations of the target variable, and we would encourage readers to use the code we share alongside this paper to apply  the approach to their own target variables, and including other auxiliary grids, to see what light Bayesian deep learning can shed on other geostatistical applications.

\subsection*{Neural network architecture}

We constructed our deep neural network in the Tensorflow framework\cite{abadi2016tensorflow}. As is shown in \autoref{fig:overview}, our network has two separate input branches: a five-layer convolutional branch that takes auxiliary variable images as input; and a single layer dense branch that takes location variables as input. The outputs of these two branches are flattened and combined into a single 1024 dimensional vector (512 from the convolutional branch, 512 from the dense branch) that feeds into the `top section' of our neural network, which consists of a further two dense layers (256 nodes, and 128 nodes) before outputting the two parameters --- mean, $\mu$, and variance, $\sigma^2$ --- of a Gaussian distribution. Throughout the network, we use the Rectified Linear Unit (`ReLU') activation function, and our loss function is the negative log-likelihood i.e. through gradient descent the neural network aims to maximise the likelihood.

In our convolutional branch, the first four layers are convolutional layers, each with 128 channels, using 3x3 kernels with dilation and stride of 1 (apart from the first, which uses a stride of 3 in order to rapidly reduce the spatial dimensions of the feature learning branch from 32x32 (the input image) to 10x10 and therefore reduce the number of parameters in subsequent convolutional layers). The fifth layer of the branch is an average pooling layer, with a 3x3 pool size and a stride of 1, which reduces the final convolutional layer's 4x4 output to a 2x2 output. Pooling is generally used as a way of instilling translation invariance into convolutional neural networks - the exact position at which filters are activated is lost during the pooling process. In the case of average pooling, the output simply reports the average activation of each filter within the pool. In our case, this reduces our convolutional output into four pools, each a 128 channel summary of learned features within each quadrant around the observation: north-west, north-east, south-west and south-east. While it may seem counter-intuitive to instill any level of translation invariance in a spatial mapping problem, we found that it helped to reduce overfitting in the network. The dense branch is a simpler affair, and follows a typical fully-connected neural network architecture. In total our neural network architecture has 741 634 trainable parameters.

\subsection*{Monte Carlo dropout}

The Gaussian distribution in the output of our neural network quantifies aleatoric uncertainty. This `natural' variability assumed in the data generating process expresses the belief that unexplained behaviour in the data is random. However, epistemic uncertainty -- uncertainty in the model itself -- is not explicitly captured. This relates to estimation uncertainty, whereby the same neural network would return different prediction if trained on data of a different size. To remedy this, we use Monte Carlo dropout to quantify this uncertainty as an approximation to Bayesian inference. For complete details of this approach, we refer to the work of Yarin Gal\cite{gal2016dropout}, but we explain its principals below.

First, consider the simpler scenario of using our neural network without dropout. In this scenario, the gradient descent training procedure aims to find a fixed set of weights, $w*$, which maximise the likelihood, i.e. maximise the probability of the data given those weights, $p(D|w*)$. Given its enormous number of parameters, our neural network could potentially achieve this by fitting the mean, $\mu$, directly through our training observations, and setting the variance, $\sigma^2$ close to zero everywhere - although this would undoubtedly be a case of overfitting the data. Regularisation techniques can be used to prevent this overfitting by penalising complexity (resulting in a smoother mean with higher variance), but regardless, the outcome would still be a fixed set of weights, $w*$.

In order to quantify epistemic uncertainty as in Bayesian inference, we do so through the posterior distribution of the weights, $p(w|D)$, given the data. This can be obtained by combining the likelihood, $p(D|w)$, with a prior distribution for the weights, $p(w)$. The prior is constructed by assuming fixed weights $\beta$ and a Bernoulli random variable $z\sim \mbox{Bern}(\pi)$ where $\pi=\mbox{Pr}(z=1)$. Then the actual weights are defined as $w=\beta z$, basically defining whether the weight $\beta$ is `active' ($z=1$) with probability $\pi$ or `inactive' ($z=0$) with probability $1-\pi$, where $1-\pi$ is the dropout rate to be tuned offline. Using Bayes rule we can obtain the posterior as:

\begin{equation}\label{bayes}
 p(w|D)=\frac{p(D|w)p(w)}{p(D)}
\end{equation}

This posterior distribution over the weights is the key to capturing epistemic uncertainty --- it represents the uncertainty in the function, or model, that the neural network has learned to approximate. Given the posterior distribution over weights, $p(w|D)$, the full posterior predictive distribution of the output $y$ given a value of the associated input $x$, is obtained from the neural network by the following integral:

\begin{equation}\label{predictive}
 p(y|x, D) = \int_{}^{}p(y|x,w)p(w|D)dw,
\end{equation}

where $p(y|x,w)$ is the assumed Gaussian distribution (whose mean and variance are functions of $x$ and $w$). In practice we approximate this integral by `Monte Carlo dropout': simulating from the posterior predictive distribution for a given input, $x$, one sample at a time, where each sample is drawn from the Gaussian distribution using a different set of weights $w$, sampled from their respective posterior distribution $p(w|D)$. As the neural network is trained with dropout active, gradient descent aims to set the weights such that the likelihood is maximised with the constraint of a fixed dropout rate. Therefore, during training the neural network converges on a set of weights that produce a posterior predictive distribution that matches the true distribution of data as closely as possible.

These equations dictate that our posterior predictive distribution is sensitive to our weights prior, $p(w)$. While not non-informative, the dropout Bernoulli prior leaves little room to impose user beliefs on the neural network - in the end it can only switch on or off weights which have been learned through likelihood-maximising gradient descent. The only user-choice comes down to setting the rate, which relates to the scale of epistemic uncertainty in the neural network, and this is treated as a hyper-parameter to be tuned according to predictive performance on an evaluation dataset. Like any other hyper-parameter, there should exist some optimal dropout rate that minimises the loss function of the predictive distribution with respect to held-out evaluation data. As in the example we considered earlier, when dropout is inactive (rate = 0), gradient descent may succeed in maximising the likelihood by overfitting the neural network, underestimating aleatoric uncertainty. At the other end of the scale, with the dropout rate too high the neural network's only chance of achieving reasonable likelihood is to underfit, and also overestimate aleatoric uncertainty. In a probabilistic context, predictive performance outside of the training data must by necessity go hand in hand with quantification of uncertainties.

\end{document}